\pdfoutput=1

\documentclass[11pt]{article}
\usepackage{ACL2023}

\usepackage{times}
\usepackage{latexsym}
\usepackage[T1]{fontenc}
\usepackage[utf8]{inputenc}
\usepackage{microtype}

\usepackage{inconsolata}
\usepackage{array}
\usepackage{times}
\usepackage{latexsym}
\usepackage{enumitem}
\usepackage{multirow}
\usepackage{booktabs}
\usepackage{amsmath}
\usepackage{arydshln}
\usepackage{graphicx}
\usepackage{graphics}
\usepackage{subfig}
\usepackage{amsfonts}
\usepackage{color}
\usepackage{url}
\usepackage{CJKutf8}

\DeclareMathOperator*{\argmax}{arg\,max}

\title{Revisiting Non-Autoregressive Translation at Scale}
\author{Zhihao Wang$^{1,3*}$, Longyue Wang$^{2*}$, Jinsong Su$^{1,3\dagger}$, Junfeng Yao$^3$, Zhaopeng Tu$^2$ \\
        $^1$School of Informatics, Xiamen University, China \\
        $^2$Tencent AI Lab, China \\        
        $^3$Key Laboratory of Digital Protection and Intelligent Processing of Intangible Cultural Heritage \\of Fujian and Taiwan (Xiamen University), Ministry of Culture and Tourism, China\\
        \texttt{zhwang@stu.xmu.edu.cn}~~~\texttt{\{vinnylywang,zptu\}@tencent.com}\\ \texttt{\{jssu,yao0010\}@xmu.edu.cn}}

\begin{document}
\maketitle
\renewcommand{\thefootnote}{\fnsymbol{footnote}}
\footnotetext[1]{Equal Contribution.}
\footnotetext[2]{Corresponding Author.}
\renewcommand{\thefootnote}{\arabic{footnote}}
\begin{abstract}
In real-world systems, scaling has been critical for improving the translation quality in autoregressive translation (AT), which however has not been well studied for non-autoregressive translation (NAT). In this work, we bridge the gap by systematically studying the impact of scaling on NAT behaviors. Extensive experiments on six WMT benchmarks over two advanced NAT models show that scaling can alleviate the commonly-cited weaknesses of NAT models, resulting in better translation performance. To reduce the side-effect of scaling on decoding speed, we empirically investigate the impact of NAT encoder and decoder on the translation performance. 
Experimental results on the large-scale WMT20 En-De show that the asymmetric architecture (e.g. bigger encoder and smaller decoder) can achieve comparable performance with the scaling model, while maintaining the superiority of decoding speed with standard NAT models. To this end, we establish a new benchmark by validating scaled NAT models on the scaled dataset, which can be regarded as a strong baseline for future works. We release code and system outputs at \url{https://github.com/DeepLearnXMU/Scaling4NAT}.
\end{abstract}

\section{Introduction}

Recent years have seen a surge of interest in non-autoregressive translation (NAT)~\cite{NAT}, which can improve the decoding efficiency by predicting all tokens independently and simultaneously. The majority studies on NAT focus on the base models trained on medium-scale datasets (e.g., Mask-Predict: 69M; WMT14 En-De: 4.5M)~\cite{ghazvininejad2019mask}, while scaled models and datasets become de facto standard for autoregressive translation (AT) models (e.g., Transformer-Big: 226M; WMT20 En-De: 45.1M)~\cite{ott2018scaling}. The model- and data-level gaps make the progress of NAT lag behind that of AT, which limits the applicability of NAT models to practical scenarios.

This general tendency motivates us to boost NAT models from the scaling perspective, including the amounts of training data and the model size. 
In this paper, we aim to provide empirical answers to the following research questions: 

\begin{itemize}[leftmargin=*,topsep=0.1em,itemsep=0.1em,parsep=0.1em]

\item {\bf RQ1}: {\em How does scaling affect NAT behaviours in terms of translation quality and decoding speed?} Scaling neural networks brings dramatic quality gains over translation tasks using AT models~\cite{arivazhagan2019massively}, and revisiting existing methods on a large-scale data can obtain more consistent conclusions~\cite{edunov2018understanding}. 

\item {\bf RQ2}: {\em Have performance improvements of scaling been accompanied by alleviating commonly-cited weaknesses of NAT?} Several weaknesses exist in NAT, including multimodality problem~\cite{NAT}, non-fluent outputs~\cite{Du:2021:ICML} and inadequate translations~\cite{Ding:2021:ICLR}. 

\item {\bf RQ3}: {\em Can we establish a new NAT benchmark to reliably translate leaderboard scores to improvements in real-world use of the models?} Although previous studies of NAT have achieved comparable performance with the AT models, they are still validated on small-scale datasets and model sizes using inconsistent evaluation criteria. These gaps make the progress of NAT lag behind that of AT, which limits the applicability of NAT models to practical scenarios.

\end{itemize}

To answer these research questions, we investigate the effects of different scaling methods on two advanced NAT models. 
Experimental results show that scaling works well with knowledge distillation to alleviate commonly-cited weaknesses of NAT. The scaled NAT models achieve better translation quality at the expense of decreasing decoding speed. To balance effectiveness and efficiency, we compare various component-scaled NAT models and find that scaling architecture in NAT is more asymmetric than that in AT. Accordingly, we introduce a cone architecture for NAT with a deeper and wider encoder and a shallower and narrower decoder, which boosts translation performance and maintain the decoding speed. 
Specifically, our {\bf main contributions} are as follows:
\begin{itemize}[leftmargin=*,topsep=0.1em,itemsep=0.1em,parsep=0.1em]
    \item We demonstrate the necessity of scaling model and data for NAT models, which narrows the progress gap between NAT and AT models.
    \item Our study reveals positive effects of scaling on commonly-cited weakness, which makes the standard NAT model sub-optimal.
    \item We establish a new benchmark, where we evaluate competing scaled NAT models on large-scale datasets in terms of effectiveness and efficiency. 
    \item We provide a better understanding of NAT at scale to help prioritize future exploration towards making NAT a common translation framework. 
\end{itemize}

\section{Preliminary}

\subsection{Non-Autoregressive Translation}

Given a source sentence ${\bf x} = \{ {x_1},{x_2}, \dots ,{x_{T_X}}\}$, an AT model generates each target word ${\bf y}_t$ conditioned on previously generated ones ${\bf y}_{<t}$, leading to high latency on the decoding stage. In contrast, NAT models break this {\em autoregressive factorization} by producing all target words independently and simultaneously. Formally, the probability of generating ${\bf y} = \{ {y_1},{y_2}, \ldots ,{y_{T_Y}}\}$ is computed as: $p({\bf y}|{\bf x})=\prod_{t=1}^{T_Y}p({\bf y}_t|{\bf x}; \theta)$
where $T_Y$ is the length of the target sequence, which is usually predicted by a separate conditional distribution. The parameters $\theta$ are trained to maximize the likelihood of a set of training examples according to $\mathcal{L}(\theta) = \argmax_{\theta} \log p({\bf y}|{\bf x}; \theta)$. 

\paragraph{Knowledge Distillation Training}

NAT suffers from the {\em multimodality problem}, where the conditional independence assumption prevents a model from properly capturing the highly multimodal distribution of target translations~\cite{NAT}. Accordingly, the sequence-level knowledge distillation~\cite{kim2016sequence} is introduced to reduce the modes of training data by replacing their original target-side samples with sentences generated by an AT teacher~\cite{NAT,zhou2019understanding}. Formally, the original parallel data $D_{\text{Raw}}$ and the distilled data $D_{\text{KD}}$ can be defined as 
$D_{\text{Raw}} = \{(\mathbf{x}_i, \mathbf{y}_i)\}^N_{i=1}$ and $D_{\text{KD}} = \{(\mathbf{x}_i, f_{s \mapsto t}(\mathbf{x}_i))|\mathbf{x}_i \in D_\text{Raw}\}^N_{i=1}$,
where $f_{s \mapsto t}$ represents an AT model trained on $D_{\text{Raw}}$ for translating sentences from the source to the target language. $N$ is the total number of sentence pairs in the training data. 

\subsection{Advanced Models}
The conditional independence assumption results in a performance gap between the NAT model and its AT teacher.
A number of recent efforts have explored ways to bridge the performance gap with advanced architectures~\cite{ghazvininejad2019mask,gu2019levenshtein,ding-etal-2020-context} or training objectives~\cite{shao2019minimizing,axe,Du:2021:ICML}. Another thread of work focuses on understanding and improving distillation training~\cite{zhou2019understanding,Ding:2021:ICLR,huang2021improving,ding2021progressive,ding2021rejuvenating,ding2022redistributing}.
Generally, NAT models can be divided into two categories:

\paragraph{Iterative NAT} is proposed to refine previously generated words in each iteration, which allows NAT models to generate target words by capturing partial and noisy dependencies. {Mask-Predict} (MaskT)~\cite{ghazvininejad2019mask} uses the conditional masked language model~\cite{devlin2019bert} to iteratively generate the target sequence from the masked input. 
{Levenshtein Transformer}~\cite{gu2019levenshtein} introduces three steps: {deletion}, {placeholder prediction} and {token prediction}, and the decoding iterations adaptively depend on certain conditions.

\paragraph{Fully NAT} is trained to produce one-pass decoding without sacrifice of speed-up. Several studies have been proposed to improve the fully NAT models~\cite{glat,gu2020fully}. GLAT adopts an adaptive glancing sampling strategy for training, which can be seen as a method of curriculum learning. Furthermore, \citeauthor{gu2020fully}~\shortcite{gu2020fully} build a new SOTA fully NAT model by combining useful techniques in four perspectives, including training data, model architecture, training objective and learning strategy.

\subsection{Experimental Setup}
\label{sec:2.3}

\paragraph{Datasets} We not only experiment on the widely-used WMT16 English-Romanian (0.6M) and WMT14 English-German (4.5M) benchmarks, but also broaden the investigation on a large-scale dataset WMT20 English-German (45.1M). 
We tokenize data using the Moses toolkit, and then split them into subwords using a joint BPE~\cite{Sennrich:BPE} with 32K merge operations. This forms a shared vocabulary of 32k, 37k, and 49k for WMT16 En-Ro, WMT14 En-De and WMT20 En-De respectively. Both AT and NAT models are trained on KD data, except as otherwise noted. To generate KD data, we employ Transformer-Big and Transformer-Base as teachers to distill the En-De and En-Ro datasets, respectively.

\paragraph{NAT Models} We validate two advanced models, representing iterative and fully NAT respectively:
\begin{itemize}[leftmargin=*,topsep=0.1em,itemsep=0.1em,parsep=0.1em]
    \item {\em MaskT}~\cite{ghazvininejad2019mask} where we follow its optimal settings to keep the iteration number be 10 and length beam be 5.
    \item {\em GLAT}~\cite{glat} where we follow their reported configurations to set iteration number and length beam as 1.
\end{itemize}
Models are re-implemented on top of the Fairseq framework~\cite{ott2019fairseq}, which supports training on multiple GPU instances. We employ {\em large-batch training} (i.e. 480K tokens/batch) to optimize the performance~\cite{ott2018scaling}. We train all NAT models for 300K steps to ensure adequate training, apart from WMT16 En-Ro (30K steps). Following the common practices~\cite{ghazvininejad2019mask,kasai2020parallel}, we evaluate the performance on an ensemble of 5 best checkpoints (ranked by validation BLEU) to avoid stochasticity. More details about NAT training are presented in Appendix~\ref{app:training}.

\paragraph{AT Teachers} 
We closely follow previous works on NAT to apply sequence-level knowledge distillation to reduce the modes of the training data. We trained \textsc{Base} and \textsc{Big} Transformer~\cite{transformer} as the {\em AT teachers} for En$\leftrightarrow$Ro and En$\leftrightarrow$De tasks, respectively. We adopt {\em large-batch training} (i.e. 458K tokens/batch) to optimize the performance of AT teachers~\cite{ott2018scaling}. Specially, the AT teachers are trained on raw data.

\paragraph{Evaluation} For fair comparison, we use case-insensitive tokenBLEU \cite{papineni2002bleu} to measure the translation quality on WMT16 En-Ro and WMT14 En-De. We use SacreBLEU \cite{post2018call} for the new benchmark WMT20 En-De.

\begin{table*}
\centering
\scalebox{0.95}{
\setlength{\tabcolsep}{4pt}
\begin{tabular}{l rr rr rr rr}
\toprule
\multirow{2}{*}{\bf Model} & \multirow{2}{*}{\bf Iter.} & \multirow{2}{*}{\bf ~~~~~~Size} & \multicolumn{2}{c}{\textbf{W16 (0.6M)}} & \multicolumn{2}{c}{\textbf{W14 (4.5M)}} & \multicolumn{2}{c}{\textbf{W20 (45.1M)} }\\
\cmidrule(lr){4-5}\cmidrule(lr){6-7}\cmidrule(lr){8-9}
& & &\textbf{En-Ro} &\textbf{Ro-En} & \textbf{En-De} &\textbf{De-En} & \textbf{En-De} &\textbf{De-En}\\
\toprule
\multicolumn{9}{c}{\bf AT Models} \\
Transformer-Base ({\em En$\leftrightarrow$Ro teacher})  & n/a &  69M & 33.9 & 34.1 & - & - & - & - \\
Transformer-Big  ({\em En$\leftrightarrow$De teacher})  & n/a & 226M & - & - & 29.2 & 32.0 & 32.4 & 41.7 \\
\toprule
\multicolumn{9}{c}{\bf Existing NAT Models} \\
AXE~\cite{axe}                                          &  1  &  69M & 30.8 & 31.5 & 23.5 & 27.9 & n/a & n/a\\
Fully-NAT~\cite{gu2020fully}                     &  1  &  70M & 33.8 & 33.9 & 27.5 & 31.1 & n/a & n/a\\
\hdashline
{DisCo}~\cite{kasai2020parallel}                        &  5  &  69M &  33.2  & 33.3 & 27.3    &  31.3& n/a & n/a\\
Imputer~\cite{imputer}                                  &  5  &  69M &  34.4  & 34.1 & 28.2    &  31.8& n/a & n/a\\
CMLMC~\cite{huang2021improving}                     & 10  &  73M & 34.6 & 34.1 & 28.4 & 31.4 & n/a & n/a\\
\toprule
\multicolumn{9}{c}{\bf Our NAT Models} \\
MaskT with {\em iterative decoding} & \multirow{4}{*}{10} &  69M & 33.9 & 33.6 & 24.7 & 29.1 & 27.2 & 36.6 \\
~~~~ + Knowledge Distillation       &                     &  69M & 34.8 & 33.8 & 27.5 & 31.1 & 31.3 & 40.6 \\
~~~~ + Width Scaling (i.e., NAT-Big)&                     & 226M & 34.6 & 33.2 & 24.9 & 29.6 & 30.2 & 38.7 \\
~~~~ + Both                         &                     & 225M & 34.7 & 34.0 & 28.2 & 31.2 & 32.9 & 42.1 \\
\hdashline
GLAT with {\em fully decoding}      & \multirow{4}{*}{1}  &  71M & 30.0 & 31.2 & 19.3 & 26.7 & 24.0 & 36.1 \\
~~~~ + Knowledge Distillation       &                     &  70M & 32.3 & 32.6 & 26.2 & 30.3 & 30.6 & 40.0 \\
~~~~ + Width Scaling (i.e., NAT-Big)&                     & 230M & 32.0 & 32.3 & 21.8 & 27.6 & 28.8 & 38.6 \\
~~~~ + Both                         &                     & 229M & 34.5 & 34.2 & 27.4 & 30.9 & 32.3 & 41.1 \\
\bottomrule
\end{tabular}}
\caption{Translation performance (BLEU $\uparrow$) of NAT models on translation tasks with different data sizes. ``Iter.'' indicates the number of iterations while ``Size'' shows the number of model parameters. ``+ Width Scaling'' denotes scaled NAT architecture similar to Transformer-Big (i.e., (6,6)$\times$1024). ``+ Knowledge Distillation'' represents training models on KD data instead of the original ones.}
\label{tab:scaled}
\end{table*}

\section{Scaling Behaviors of NAT Models}

We investigate effects of scaling from three perspectives: translation quality, commonly-cited weaknesses and decoding efficiency. The settings are:
\begin{itemize}[leftmargin=*,topsep=0.1em,itemsep=0.1em,parsep=0.1em]
    \item {\em Model Scaling}: Based on traditional NAT-Base configurations ((6,6)$\times$512), we conduct 1) Width Scaling, where the size of feed-forward dimensions are enlarged to 1024 (NAT-Big: (6,6)$\times$1024); 2) Depth Scaling, where the number of stacked encoder-decoder layers is increased to to 24-24 (NAT-Deep: (24,24)$\times$512)). We mainly investigate behaviors of NAT-Big as it has similar performance with NAT-Deep, while the training of NAT-Big is more stable.
    
    \item {\em Data Scaling}: The commonly-used datasets for NAT are WMT16 En-Ro and WMT14 En-De, whose sizes are smaller than current AT benchmarks. We mainly experiment NAT models on the WMT20 En-De dataset, which is 10 times larger than previous ones (i.e. WMT16: 0.6M; WMT14: 4.5M; WMT20 45.1M).
\end{itemize}

\subsection{Translation Quality}

\paragraph{Results on Benchmarks}
Table~\ref{tab:scaled} lists the results on the six benchmarks: WMT16 En$\leftrightarrow$Ro, WMT14 En$\leftrightarrow$De, and WMT20 En$\leftrightarrow$De, which are small-, medium- and large-scale datasets, respectively. We experiment MaskT and GLAT models of whose configurations are detailed in Section~\ref{sec:2.3}.
Compared with standard NAT models (``+ Knowledge Distillation''), scaling method (``+ Both'') significantly and consistently improves translation performance (BLEU$\uparrow$) on medium and large-scale datasets. However, the improvement is not robust on small-scale dataset. 
An interesting finding is that both model scaling and data scaling are able to narrow the performance gap between fully and iterative NAT models. After model scaling, the average difference between MaskT and GLAT drops from +1.2 (``+ Knowledge Distillation'' lines) to +0.5 (``+ Both'' lines). 
Encouragingly, advanced NAT models with model-scaling can perform better than strong AT teachers on larger-scale data. As seen, the performance of ``MaskT+Both'' is +0.5 higher than the Transformer-Big models on WMT20 En$\leftrightarrow$De.
{\em This confirms the necessarily of scaling model size and data for building practical and robust NAT systems.}

\paragraph{Complementary between Scaling and KD}
KD is a commonly-used training recipe to boost NAT performance. As shown in Table~\ref{tab:scaled}, KD (``+ Knowledge Distillation'') can benefit more for fully NAT than iterative NAT models compared with Raw (+4.1 vs. +2.3 BLEU scores averagely). 
We also find that KD is more effective on large-scale datasets, where the average improvements are +4.7 and +2.5 on WMT20 and WMT16+14, respectively. This reconfirms the effectiveness of KD training especially on large data.
The model scaling (``+ Width Scaling'') can also improve NAT models by enhancing the model ability on learning difficult data. The conclusions of model scaling are similar to KD: 1) it benefits more for fully NAT (+1.0 vs. +2.2 BLEU); 2) it is more effective on large-scale datasets (+3.0 vs. +0.9 BLEU). 
Combining scaling with KD (``+ Both'') can further improve standard MaskT and GLAT (``+ Knowledge Distillation'') by +0.7 and +1.3, which illustrates that they exhibit complementary properties for NAT models. 
We extensively analyze the reasons behind this in Section~\ref{sec:3.2}.
{\em Scaling and KD are related to and complement one another for NAT models.} 
The conclusion on complementary between scaling and KD also holds for depth scaling (detailed in Appendix \S\ref{app:deep_scaling}). 
The deep models also have similar performance with big ones, but depth scaling is difficult to train with side effect on inference speed. Therefore, we employ NAT-Big as our testbed in following experiments, unless otherwise specified.

\subsection{Difference between NAT and AT Scaling}

The scaling behavior of AT models has been studied~\cite{wang2019learning}, which seems similar to NAT in terms of BLEU score. Different from autoregressive Transformer, NAT predicts target tokens independently and simultaneously, which may lead to different scaling behaviors of NAT models. Starting from this intuition, we further compare NAT and AT scaling from the perspective of linguistic properties. Probing tasks~\cite{probingtask} can quantitatively measure the linguistic knowledge embedded in encoder representations. We follow \citeauthor{hao2020multi}~\shortcite{hao2020multi} to analyze Base and Big models trained on WMT20 En$\rightarrow$De KD data. The experimental results on WMT20 En$\rightarrow$De raw data are also provided in Appendix \S\ref{sec:probing_task_for_raw}.

As depicted in Table~\ref{tab:probing}, scaling improves NAT and AT models on syntatic (+1.4\% vs. +1.7\%, averagely) and semantic (+0.7\% vs. +1.0\%, averagely) abilities. However, their behaviors are quite different on surface tasks (-0.1\% vs. +12.9\%, averagely), which test the ability of preserving global information contained in sentence embeddings.
Specifically, scaling improves ability of AT models on ``WC'' subtask (+18.4\%), while this weakens NAT ability (-3.5\%). Besides, NAT-Base model preserves more surface information than AT-Base (SeLen: 80.6\% vs. 78.1\%; WC: 81.3\% vs. 55.6\%).

\begin{table}
\centering
\setlength{\tabcolsep}{4pt}
\scalebox{0.95}{
\begin{tabular}{l r rrr rrr}
\toprule
\multicolumn{2}{c}{\multirow{2}{*}{\bf Task}} & \multicolumn{3}{c}{\bf{MaskT}} & \multicolumn{3}{c}{\bf{AT}} \\
\cmidrule(lr){3-5}\cmidrule(lr){6-8}
& & \bf{Base} & \bf{Big} & $\Delta$ & \bf{Base} & \bf{Big} & $\Delta$ \\
\midrule
\multirow{2}{*}{\rotatebox[origin=c]{90}{\small Surface}} & SeLen & 80.6 & 83.9 & +3.3 & 78.1 & 85.4 & +7.3\\
& WC & 81.3 & 77.8 & {\bf -3.5} & 55.6 & 74.0 & {\bf +18.4} \\
\midrule
\multirow{3}{*}{\rotatebox[origin=c]{90}{\small Syntactic}} & TrDep & 35.2 & 36.9 & +1.7 & 35.8 & 36.9 & +1.1\\
& ToCo & 70.8 & 73.0 & +2.2 & 69.0 & 72.9 & +3.9\\
& Bshif & 49.6 & 49.9 & +0.3 & 50.1 & 50.1 & 0\\
\midrule
\multirow{5}{*}{\rotatebox[origin=c]{90}{\small Semantic}} & Tense & 83.8 & 85.1 & +1.3 & 84.4 & 85.5 & +1.1\\
& SubN & 79.7 & 80.9 & +1.2 & 79.7 & 80.0 & +0.3\\
& ObjN & 81.5 & 82.0 & +0.5 & 80.6 & 82.1 & +1.5\\
& SoMo & 49.9 & 49.9 & 0 & 49.7 & 49.9 & +0.2\\
& CoIn & 53.4 & 53.9 & +0.5 & 53.0 & 55.0 & +2.0\\
\bottomrule
\end{tabular}}
\caption{Performance (accuracy $\uparrow$) of probing tasks for evaluating linguistic properties embedded in the learned representations of AT and NAT models (Width Scaling).}
\label{tab:probing}
\end{table}

\subsection{Analysis on NAT Weaknesses}
\label{sec:3.2}

We analyze effects of scaling on commonly-cited weaknesses: 1) {\em multimodality} indicated by token repetition ratio~\cite{NAT}; 2) {\em generation fluency} calculated by language model (LM) perplexity~\cite{Du:2021:ICML}; 3) {\em translation adequacy} measured by word translation accuracy~\cite{Ding:2021:ICLR}. Table~\ref{tab:analysis} shows the results. Examples about NAT weaknesses are listed in Appendix~\ref{app:weakness}.

\paragraph{Scaling Alleviates Multimodality Problem} 

Repeated token percentage is a commonly-used metric of measuring multimodality in a NAT model~\cite{imputer}. A NAT model may consider many possible translations at the same time due to the independent predictions of target tokens. Accordingly, the NAT output typically contains some repetitive tokens, especially for fully NAT (1.1\% vs. 2.7\%). Similar to KD, scaling is an alternative method to significantly reduce the repetition percentage for NAT models (-0.5\% and -1.0\%). In addition, combining KD and scaling can further alleviate the repetition problem, which is consistent with the translation quality in Table~\ref{tab:scaled}.

\begin{table}
\centering
\scalebox{0.96}{
\setlength{\tabcolsep}{2.8pt}
\begin{tabular}{l rr rr rr}
\toprule
\multirow{2}{*}{\bf Model}   &   \multicolumn{2}{c}{\bf Repetition $\downarrow$ } &  \multicolumn{2}{c}{\bf PPL  $\downarrow$}   &   \multicolumn{2}{c}{\bf WA $\uparrow$}\\
\cmidrule(lr){2-3}\cmidrule(lr){4-5}\cmidrule(lr){6-7}
& \bf \# & \bf $\Delta$ &  \bf \#  & \bf $\Delta$ & \bf \# & \bf $\Delta$\\
\midrule
MaskT         & 1.1\% & -      & 66 & -   & 71.3\% & - \\
~~~~ + KD     & 0.2\% & -0.9\% & 55 & -11 & 73.0\% & +1.7\%\\
~~~~ + Scale  & 0.6\% & -0.5\% & 59 & -7  & 72.2\% & +0.9\%\\
~~~~ + Both   & 0.1\% & -1.0\% & 52 & -14 & 73.4\% & +2.1\%\\
\midrule
GLAT          & 2.7\% & -      & 98 & -   & 70.7\% & - \\
~~~~ + KD     & 1.2\% & -1.5\% & 70 & -28 & 72.6\% & +1.9\%\\
~~~~ + Scale  & 1.7\% & -1.0\% & 79 & -19 & 72.1\% & +1.4\%\\
~~~~ + Both   & 0.8\% & -1.9\% & 64 & -34 & 73.1\% & +2.4\%\\
\midrule
\bf Golden &  0.02\%  & -      & 54 & -   & -      & - \\
\bottomrule
\end{tabular}}
\caption{Analyses of translation outputs generated by NAT models on the WMT20 De$\rightarrow$En test set. Lower repeated token percentages (``Repetition") represent lower multimodality in a model. Lower perplexities (``PPL") denote better fluency while higher word translation accuracy (``WA'') denotes better adequacy. \# is the absolute value and $\Delta$ is the difference over NAT-Base models. We also list the results of reference (``Golden'').} 
\label{tab:analysis}
\end{table}

\paragraph{Scaling Improves Generation Fluency} 

NAT models typically suffer from fluency problems because they only have limited capabilities to model dependencies between the target tokens~\cite{kasner2020improving,gu2020fully}.
We measure the fluency of output with a public released LM,\footnote{\url{https://github.com/pytorch/fairseq/tree/main/examples/wmt19}.} which is trained on the News Crawl corpus. 
The results show that either KD or scaling can consistently decrease the PPL in all cases (-7$\sim${-34}). We attribute the improvement of fluency to that KD reduces the learning difficulty by simplifying training data while scaling enhances the model ability by introducing larger parameters. Besides, the complementarity between KD and scaling still holds in terms of fluency measurement.
Encouragingly, scaled model without KD performs closely to the standard NAT models, showing that scaling has the potential to directly learn from the raw data of complex modes. 

\paragraph{Scaling Enhances Translation Adequacy} 

NAT often suffers from two kinds of adequacy errors which were empirically observed by previous studies: 1) incomplete translation, due to incomplete transfer of source side information~\cite{natreg}; (2) lexical choice, due to choosing a target lexeme which inadequately expresses the source meaning~\cite{Ding:2021:ICLR}. Following  \citeauthor{compare-mt}~\shortcite{compare-mt}, we measure the word accuracy which defined as F-measure of system outputs with respect to the reference. It can demonstrate how much a system over- or under-produces words of a specific type as well. As expected, NAT models with KD or scaling have higher word accuracy (+0.9\%$\sim$+1.9\%), resulting in better translation performance (BLEU$\uparrow$ in Table~\ref{tab:scaled}). Combing KD and scaling can further improve translation quality by increasing word accuracy (+2.1\%$\sim$+2.4\%).

\subsection{Discussion on Decoding Efficiency}

\begin{table}
\centering
\scalebox{0.95}{
\setlength{\tabcolsep}{2.4pt}
\begin{tabular}{l r rc rc r}
\toprule
\multirow{2}{*}{\bf Model} & \multirow{2}{*}{\bf Size} & \multicolumn{2}{c}{\textbf{Speed$_\text{1}$}} & \multicolumn{2}{c}{\textbf{Speed$_\text{max}$}} & \multirow{2}{*}{\bf BLEU}\\
\cmidrule(lr){3-4}\cmidrule(lr){5-6}
& & \# & $\Delta$ & \# & $\Delta$\\
\midrule
MaskT        &  69M &  8.9  &  -           &  166 & - & 31.3\\
~~ + Scale   & 225M &  8.4 &  $0.94\times$ &   92 & $0.55\times$ & 32.9\\
\hdashline
GLAT         &  70M & 58.8 & -            & 2160 & - & 30.6\\
~~ + Scale   & 229M & 54.4 & $0.93\times$ & 1772 & $0.82\times$ & 32.3\\
\bottomrule
\end{tabular}}
\caption{Decoding speed (sentences/s $\uparrow$) of scaled NAT models (i.e., NAT-Big (6, 6)$\times$1024) on the WMT20 En$\rightarrow$De task. We test the decoding speed when translating one sentence (Speed$_\text{1}$) or hardware-maximum mini-batches (Speed$_\text{max}$). \# is the absolute value and $\Delta$ is the speedup ratio over NAT-Base.}
\label{tab:speed}
\end{table}

Although scaling produces significant performance gains, someone may argue that model scaling introduces more parameters, which will increase latency at decoding stage. Following previous studies, we carefully investigate effects of scaling on decoding efficiency for NAT. We employ two metrics:
\begin{itemize}[leftmargin=*,topsep=0.1em,itemsep=0.1em,parsep=0.1em]
\item {\em Speed$_\text{1}$}, which measures speed when translating one sentence at a time~\cite{NAT}. This is used in standard practice and aligns with applications like instantaneous MT that translates text input from users immediately.
\item {\em Speed$_\text{max}$}, which measures speed when translating in mini-batches as large as the hardware allows~\cite{kasai2021deep}. This corresponds to scenarios where one wants to translate a large amount of text given in advance.
\end{itemize}

As illustrated in Table~\ref{tab:speed}, adding 3$\times$ parameters definitely decreases the decoding speed (Speed$_\text{1}$: $0.93\times\sim0.94\times$ and Speed$_\text{max}$: $0.55\times\sim0.82\times$). In terms of Speed$_\text{max}$, scaling harms the iterative NAT more than fully NAT models ($0.55\times$ vs. $0.82\times$). Besides, we test the decoding speed of the MaskT-Deep model ((24, 24)$\times$512) and find that Speed$_\text{1}$ rapidly declines to $0.28\times$. These results suggest that scaling method increases translation quality (BLEU $\uparrow$) at the expense of decoding speed (Speed $\downarrow$), especially on Speed$_\text{max}$. 

This findings motivates us to design a better scaling architecture for NAT, taking both performance and time cost into consideration. \citeauthor{kasai2021deep}~\shortcite{kasai2021deep} pointed out that some NAT models have little advantage when translating a large amount of text given in advance. Accordingly, we use Speed$_\text{max}$ as default when discussing translation speed.

\section{New NAT Benchmark}
\label{sec:4}

Most NAT models are implemented upon the encoder-decoder framework, where the encoder summarizes the source sentence and the decoder learns to generate target words. We ask: {\em how to scale this framework?} In this section, we empirically search for a better NAT architecture by considering both effectiveness and efficiency.

\begin{table}
\centering
\scalebox{0.94}{
\setlength{\tabcolsep}{2.5pt}
\begin{tabular}{l ccc ccc}
\toprule
\multirow{2}{*}{\bf Model} & \multicolumn{3}{c}{\textbf{Speed$_\text{max}$}} & \multicolumn{3}{c}{\bf BLEU} \\
\cmidrule(lr){2-4}\cmidrule(lr){5-7}
& AT & MaskT & GLAT & AT & MaskT & GLAT\\
\midrule
\multicolumn{7}{c}{\bf No Scaling}\\
{Base} & $1.00\times$ & $1.00\times$ & $1.00\times$ & 33.0 & 31.3 & 30.6 \\
\midrule
\multicolumn{7}{c}{\bf Component Scaling}\\
{Enc.} & $0.99\times$ & $0.96\times$ & $0.89\times$ & 33.5 & 33.1 & 32.0 \\
{Dec.} & $0.74\times$ & $0.56\times$ & $0.85\times$ & 33.0 & 32.5 & 31.0 \\
{Both} & $0.72\times$ & $0.55\times$ & $0.82\times$ & 34.0 & 32.9 & 32.3 \\
\bottomrule
\end{tabular}}
\caption{Translation performance (BLEU $\uparrow$) and relative decoding speed of component-level width-scaled NAT models on the WMT20 En$\rightarrow$De. Speed$_\text{max}$ means speedup ratio of scaled NAT models over Base ones.}
\label{tab:scale_component}
\end{table}

\subsection{Discussion on Architecture Symmetry}
\label{sec:4.1}

Previous studies usually propose asymmetric architectures for AT such as the one with deep encoder and shallow decoder~\cite{kasai2021deep}. The main reason is that increasing the number of layers, especially in the decoder, deteriorates the latency of translation and memory costs. We verify the architecture symmetry of NAT models by investigating impacts of component-level scaling on translation quality and decoding speed. More specifically, we enlarge the size of layer dimensions in either encoder or decoder, or both components. Table~\ref{tab:scale_component} shows results of component-level width-scaling on the WMT20 En-De dataset. Results of component-level depth-scaling are shown in Appendix \S\ref{app:deeper_component}.

\paragraph{Translation Performance} 

Clearly the scaling approach improves the translation quality in all cases, although there are still considerable differences among the variants (``Component Scaling'' vs. ``No Scaling''). 
Introducing encoder- and decoder-scaling individually improves translation performance over the standard MaskT by +1.8 and +1.2 BLEU points respectively. As seen, scaling encoder and decoder are not equivalent in terms of translation performance. This asymmetric phenomenon is more severe than that in AT models. The possible reason is that NAT model need to spend a substantial amount of its capacity in disambiguating source and target words under the conditional independence assumption. 
However scaling both encoder and decoder cannot always achieve better performance compared with individual scaling. This is opposite to AT models, which can further increase by +0.5 BLEU point. {\em To sum up, 1) scaling NAT is more asymmetric than AT; 2) complementary between encoder and decoder in NAT is weaker than that in AT.}

\paragraph{Decoding Efficiency} 

Compared with Base models, scaling encoder has minimal side-effect on the decoding speed (MaskT: 0.96$\times$; GLAT: 0.89$\times$). The conclusion still holds on AT models (0.99$\times$). However, scaling decoder has a large impact on decoding speed (MaskT: 1.00$\times$ $\rightarrow$ 0.56$\times$; GLAT: 1.00$\times$ $\rightarrow$ 0.85$\times$). It is worth noting that iterative NAT is more sensitive to decoder-scaling than fully NAT. The main reason is that iterative mechanism should occupy many times of GPU memory, resulting in smaller mini-batches when calculating Speed$_\text{max}$. Furthermore, there is no further speed decrease when scaling both encoder and decoder components (MaskT: 0.56$\times$ $\rightarrow$ 0.55$\times$; GLAT: 0.85$\times$ $\rightarrow$ 0.82$\times$). {\em To sum up, 1) The decoding latency is mainly attributed to scaling decoder; 2) Scaling decoder of iterative NAT comes at the cost of a much larger time cost than fully NAT.}

\begin{table}
\centering
\scalebox{0.95}{
\begin{tabular}{c rrrr}
\toprule
{\bf Model} & {\bf Base} & {\bf Enc.} & {\bf Dec.} & {\bf Both}\\
\midrule
MaskT & 81.3 & \bf 92.4 & 85.2 & 77.8 \\
AT & 55.6 & \bf 93.0 & 87.3 & 74.0 \\
\bottomrule
\end{tabular}}
\caption{Performance (accuracy $\uparrow$) of ``WC (word-content)'' probing of component-level width-scaled NAT and AT models.}
\label{tab:component-probing}
\end{table}

\paragraph{Linguistic Probing} 

As discussed in Section~\ref{sec:3.2}, NAT and AT models have different scaling behaviors on learning word-content linguistics. We further investigate the effects at component level in Table~\ref{tab:component-probing}. {\em To sum up, asymmetric scaling can enhance the capability of NAT on learning word-content knowledge. The conclusion still holds on AT.}

\subsection{Asymmetric Scaling Method}

To find a better scaling architecture, we conduct ablation study on a variety of scaled NAT models. Based on the findings, we propose a new NAT architecture to boost translation quality without increasing latency during inference.

\begin{table}[t]
\centering
\scalebox{0.96}{
\setlength{\tabcolsep}{4pt}
\begin{tabular}{c rr rr rr r cc}
\toprule
\multirow{2}{*}{\bf \#} & \multicolumn{2}{c}{\bf Encoder} & \multicolumn{2}{c}{\bf Decoder} & \multirow{2}{*}{\bf Size} & \multicolumn{2}{c}{\bf Performance}\\
\cmidrule(lr){2-3}\cmidrule(lr){4-5}\cmidrule(lr){7-8}
& \textbf{\#L} & \textbf{Dim.} & \textbf{\#L} & \textbf{Dim.} & & \textbf{Speed} & \textbf{BLEU}\\
\midrule
1 &  6 &  512 & 6 & 512 &  69M & $1.00\times$ & 42.0\\
\midrule
2 &  6 & 1024 & 6 & 512 & 170M & $0.96\times$ & 42.6\\
3 & 12 &  512 & 6 & 512 & 105M & $0.99\times$ & 42.4\\
4 & 12 & 1024 & 6 & 512 & 246M & $0.95\times$ & 43.1\\
5 & 12 & 1024 & 6 & 256 & 217M & $1.32\times$ & 43.1\\
\midrule
6 & 12 & 1024 & 3 & 512 & 231M & $1.29\times$ & 42.7\\
7 & 12 & 1024 & 3 & 256 & 213M & $1.58\times$ & 43.0\\
\bottomrule
\end{tabular}}
\caption{Ablation Study of different NAT architectures varying from scaling methods (i.e. number of stacked layer and dimension of feed-forward) and scaling components (i.e. encoder and decoder). ``Speed'' shows the speedup ratio over NAT-Base, where we measure the decoding speed in terms of {S$_\text{max}$}. We evaluate MaskT on WMT20 En$\rightarrow$De validation set. }
\label{tab:architecure_search}
\end{table}

\begin{table*}
\centering
\scalebox{0.95}{
\setlength{\tabcolsep}{6pt}
\begin{tabular}{l rr rr rr rr}
\toprule
\multirow{2}{*}{\bf Model} & \multirow{2}{*}{\bf Speed} & \multirow{2}{*}{\bf Size} & \multicolumn{2}{c}{\textbf{W16 (0.6M)} } & \multicolumn{2}{c}{\textbf{W14 (4.5M)} } & \multicolumn{2}{c}{\textbf{W20 (45.1M)} }\\
\cmidrule(lr){4-5}\cmidrule(lr){6-7}\cmidrule(lr){8-9}
& & & \textbf{En-Ro} &\textbf{Ro-En} & \textbf{En-De} &\textbf{De-En} & \textbf{En-De} &\textbf{De-En}\\
\midrule
\multicolumn{9}{c}{\bf AT Models} \\
AT-Base     & n/a &  69M & 33.9 & 34.1 & - & - & - & - \\
AT-Big      & n/a & 226M & - & - & 29.2 & 32.0 & 32.4 & 41.7 \\
\midrule
\multicolumn{9}{c}{\bf NAT Models} \\
MaskT-Base  &  $1.00\times$ &  69M & 34.8 & 33.8 & 27.5 & 31.1 & 31.3 & 40.6 \\
MaskT-Big   &  $0.55\times$ & 225M & 34.7 & 34.0 & 28.2 & \bf 31.2 & 32.9 & \bf 42.1 \\
\hdashline
MaskT-Cone  &  $1.58\times$ & 213M & \bf 35.0 & \bf 34.5 & \bf 28.4 & 31.1 & \bf 33.2 & \bf 42.1 \\
\midrule
GLAT-Base   &  $1.00\times$ &  70M & 32.3 & 32.6 & 26.2 & 30.3 & 30.6 & 40.0 \\
GLAT-Big    &  $0.82\times$ & 229M & \bf 34.5 & \bf 34.2 & 27.4 & 30.9 & 32.3 & 41.1 \\
\hdashline
GLAT-Cone   & $0.90\times$ & 215M & \bf 34.5 & \bf 34.2 & \bf 27.7 & \bf 31.1 & \bf 32.5 & \bf 41.2  \\
\bottomrule
\end{tabular}}
\caption{Translation performance (BLEU $\uparrow$) of the proposed NAT models on translation tasks with different data sizes. ``Cone'' denotes scaling NAT architecture to (12$\times$1024, 3$\times$256). ``Speed'' shows the speedup ratio over NAT-Base, where we measure the decoding speed in terms of {S$_\text{max}$}. }
\label{tab:benchmark}
\end{table*}

\paragraph{Ablation Study} 

Seven MaskT models with different architectures are investigated on WMT20 En$\rightarrow$De dataset. These models are varying from scaling methods (i.e. depth and width) and scaling components (i.e. encoder and decoder). Table~\ref{tab:architecure_search} shows the variant configurations and the corresponding performances in terms of decoding speedup and translation quality. The \#1 is an NAT-Base model, which contains 6 encoder layers and 6 decoder layers with feed-forward dimensions being 512 (i.e. (6, 6)$\times$512). As shown in \#2$\sim$4, widening or deepening the encoder component can boost translation quality (BLEU $\uparrow$) with decreasing the decoding efficiency lightly (Speed $\downarrow$). Compared with the best encoder-scaling architecture (\#4), further widening the decoder counterpart (\#5) fails to increase the BLEU scores (43.1 vs. 43.1) but decrease the decoding speed (0.95$\times$ vs. 1.32$\times$). To better trade off efficiency and effectiveness, we make the decoder shallower and smaller based on the \#4 model. Encouragingly, the \#6 and \#7 models still achieve  comparable translation quality while increasing the speed of decoding to some extent (42.7 vs. 43.0 BLEU and 1.29$\times$ vs. 1.58$\times$ Speed). This confirms our hypothesis that NAT models need an asymmetric framework when considering both translation quality and decoding speed.

\paragraph{Cone Scaling}

Motivated by the ablation study, we propose a ``Cone'' architecture for NAT, of which encoder is deep and big while the decoder is shallow and small (i.e. (12$\times$1024, 3$\times$256)). 
As shown in Table~\ref{tab:benchmark}, we adapt the cone-scaling to {MaskT} and GLAT models, and evaluate them on {six benchmarks}. 
In general, our method achieves comparable performance with big models while retaining low latency during inference. As seen, the cone-scaling improve standard MaskT model by {+0.9} BLEU averagely) and 1.58$\times$ decoding speedup (over MaskT-Big by {+0.2} BLEU and 2.87$\times$ {Speed}).
{Besides, the cone-scaling improves standard GLAT model by +1.5 BLEU but decreases decoding speed by -0.90$\times$ (over GLAT-Big by +0.1 BLEU and 1.10$\times$ Speed).}
Surprisingly, our method can further benefit the translation quality, leading to much better performance than AT teachers (MaskT: {+0.2 BLEU averagely}). This emphasizes the need for scaling NAT as a standard procedure.
This can be used as a new benchmark over NAT models to convey the extent of the challenges they pose. 
We also measure translation quality with METEOR~\cite{meteor}, which incorporates semantic information by calculating either exact match, stem match, or synonymy match. As shown in Table~\ref{tab:meteorwmt20ende}, the cone scaling consistently achieves the best performance. 
Results on more datasets are listed in Appendix \S\ref{app:meteor}.

\section{Conclusion and Future Work}

In this study we target bridging the gap of model and data scale between NAT and AT models by investigating the scaling behaviors of NAT models. We find that simply scaling NAT models (NAT-Big) can significantly improve translation performance, especially on large-scale training data.
To better balance effectiveness and efficiency, we empirically study the contributions of scaling encoder and scaling decoder, and find that scaling NAT is more asymmetric than AT. Based on the observations, we design a new scaling architecture with deeper and wider encoder and shallower and narrower decoder (NAT-Cone), which achieves comparable performance with NAT-Big without scarifying decoding speed.
Our study empirically indicates the potential to make NAT a practical translation system as its AT counterpart.

\begin{table}[t]
\centering
\scalebox{0.95}{
\begin{tabular}{l rr rr}
\toprule
\multirow{2}{*}{\bf Model} & \multicolumn{2}{c}{\bf MaskT} & \multicolumn{2}{c}{\bf GLAT}\\
\cmidrule(lr){2-3} \cmidrule(lr){4-5}
 & En-De & De-En & En-De & De-En \\
\midrule
Base & 45.1 & 34.5 & 44.4 & 34.0 \\
Big  & 46.3 & 34.9 & 45.9 & \bf 34.5 \\
Cone & \bf 46.7 & \bf 35.0 & \bf 46.2 & \bf 34.5 \\
\bottomrule
\end{tabular}}
\caption{Translation quality of proposed NAT models in terms of METEOR ($\uparrow$) on WMT20 En$\leftrightarrow$De tasks.}
\label{tab:meteorwmt20ende}
\end{table}

However, the SOTA NAT models (including Scaling NAT) still rely on the distillation by an AT teacher. Future work will investigate better techniques to train scaled NAT models from scratch (i.e. without distillation). We additionally experiment larger NAT models in Appendix \S\ref{sec:larger_model}, which can be regarded as a preliminary experiments for this. 
We will also explore scaling NAG models in other NLP tasks, such as keyphrase generation~\cite{wrone2set} and text-to-table generation~\cite{seq2seqset}. The advent of large language models (LLMs) like GPT-4 has ushered in a new era in MT~\cite{lyu2023new,Jiao2023ParroTTD,Jiao2023IsCA,wang2023document,he2023exploring}. This innovation is causing us to reconsider conventional paradigms, especially with regards to NAT models.

\section*{Limitations}

We list the main limitations of this work as follows:
\begin{itemize}[leftmargin=*,topsep=0.1em,itemsep=0.1em,parsep=0.1em]
    \item {\bf Limited NAT Models}. The conclusions in this paper are drawn from two representative NAT models, which may be not necessarily well suited for other NAT models. The main reason is that experiments on six WMT benchmarks have cost a large number of GPU resources. We therefore appeal to future works compare more NAT models using the new benchmarks.
    
    \item {\bf Carbon Emissions}. This work totally costed 40,000 GPU hours (around 8,160 kg of CO$_2$), because 1) large numbers of experiments; and 2) scaled neural networks and training data require more GPU resources. However, we hope our empirical results can help other researchers to reduce the expense of redundant model training.
    
\end{itemize}

\section*{Ethics Statement}
We take ethical considerations very seriously, and strictly adhere to the ACL Ethics Policy.
This paper focuses on empirical evaluations on large-scale datasets and scaled NAT models, which can be seen as a reality check. 
Both the datasets and models used in this paper publicly available and have been widely adopted by studies of machine translation. We ensure that the findings and conclusions of this paper are reported accurately and objectively.

\section*{Acknowledgements}
The project was supported by National Key Research and Development Program of China(No. 2020AAA0108004), National Natural Science Foundation of China (No. 62276219), and Natural Science Foundation of Fujian Province of China (No. 2020J06001). We also thank the reviewers for their insightful comments.

\bibliography{acl2023}
\bibliographystyle{acl_natbib}

\clearpage
\appendix

\section{Appendix}
\label{sec:appendix}

\subsection{Results of Depth Scaling}
\label{app:deep_scaling}

\paragraph{Main Results}
We also exploit impacts of the depth scaling on NAT performance. Table~\ref{tab:deeper} shows the results of MaskT and GLAT models on WMT20 En-De. In general, most of conclusions in width scaling still hold for depth scaling.
Furthermore, deep and big models achieve comparable performances on KD data (MaskT: 33.1 vs. 32.9 and GLAT: 33.0 vs. 32.3 on En$\rightarrow$De; MaskT: 41.9 vs. 42.1 and GLAT: 41.2 vs. 41.1 on De$\rightarrow$En) using a comparable number of parameters (MaskT: 201M vs. 225M and GLAT: 203M vs.229M). On raw data, the performance gap between fully NAT (GLAT) and iterative NAT (MaskT) can be completely overcome by depth scaling (NAT-Base: 24.0 vs. 27.0 and NAT-Deep: 31.3 vs. 30.2 on En$\rightarrow$De; NAT-Base: 36.1 vs. 36.6 and NAT-Deep: 39.8 vs. 39.3 on De$\rightarrow$En), while width scaling can only bridge the gap between fully NAT and iterative NAT (NAT-Base: 24.0 vs. 27.0 and NAT-Big: 28.8 vs. 30.2 on En$\rightarrow$De; NAT-Base: 36.1 vs. 36.6 and NAT-Big: 38.6 vs. 38.7 on De$\rightarrow$En). It indicates that depth scaling is a better way to improve the performance of fully NAT than width scaling.

\begin{table}[h]
\centering
\setlength{\tabcolsep}{6.5pt}
\scalebox{0.95}{
\begin{tabular}{l r rr}
\toprule
\multirow{2}{*}{\bf Model} & \multirow{2}{*}{\bf Size} & \multicolumn{2}{c}{\bf WMT20}\\
\cmidrule(lr){3-4}
&     &   \bf En-De & \bf De-En\\
\midrule
MaskT                       &  69M & 27.2 & 36.6 \\
~~ + Distillation &  69M & 31.3 & 40.6 \\
~~ + Depth Scaling          & 202M & 30.2 & 39.3 \\
~~ + Both                   & 201M & 33.1 & 41.9 \\
\midrule
GLAT                        &  71M & 24.0 & 36.1 \\
~~ + Distillation &  70M & 30.6 & 40.0 \\
~~ + Depth Scaling          & 203M & 31.3 & 39.8 \\
~~ + Both                   & 203M & 33.0 & 41.2 \\
\bottomrule
\end{tabular}}
\caption{Translation performance (BLEU $\uparrow$) of NAT-Deep (``Depth Scaling'') models on WMT20 En$\leftrightarrow$De tasks. The size of NAT-Deep model is (24, 24)$\times$512, which has comparable parameters (202M vs. 226M) with the NAT-Big model in Table~\ref{tab:scaled}.}
\label{tab:deeper}
\end{table}

\begin{table}[t]
\centering
\setlength{\tabcolsep}{8pt}
\scalebox{0.95}{
\begin{tabular}{l r r}
\toprule
{\bf Model} & {\bf Size} & {\bf BLEU}\\
\midrule
MaskT-Deep (54, 54)                   & 422M & 31.4 \\
~~+ Distillation                      & 422M & 33.1 \\
\midrule
GLAT-Deep (54, 54)                    & 424M & 32.2 \\
~~+ Distillation                      & 423M & 33.3 \\
\bottomrule
\end{tabular}}
\caption{Translation performance of NAT-Deep (``Depth Scaling'') models on WMT20 En$\rightarrow$De task. The size of NAT-Deep model is (54, 54)$\times$512. Relevant experimental results of NAT-Base are shown in Table~\ref{tab:deeper}.}
\label{tab:deeper-54}
\end{table}

\begin{table}[t]
\centering
\setlength{\tabcolsep}{9pt}
\scalebox{0.95}{
\begin{tabular}{l r rr}
\toprule
\multirow{2}{*}{\bf Model} & \multirow{2}{*}{\bf Size} & \multicolumn{2}{c}{\bf BLEU}\\
\cmidrule(lr){3-4}
 & & Raw & KD \\
\midrule
MaskT                                         &  69M & 27.2 & 31.3 \\
~~ + Deep Enc.                                & 221M & 30.8 & 33.9 \\
~~ + Deep Dec.                                & 271M & 29.2 & 32.7 \\
~~ + Deep Both                                & 422M & 31.4 & 33.1 \\
\midrule
GLAT                                          &  71M & 24.0 & 30.6 \\
~~ + Deep Enc.                                & 222M & 29.1 & 32.8 \\
~~ + Deep Dec.                                & 273M & 27.4 & 31.2 \\
~~ + Deep Both                                & 424M & 32.2 & 33.3 \\
\bottomrule
\end{tabular}}
\caption{Translation performance of component-depth-scaled NAT models on WMT20 En$\rightarrow$De task. The size of NAT model is (54, 6)$\times$512 or (6, 54)$\times$512.}
\label{tab:scale_component_deeper-54}
\end{table}

\begin{table}[t]
\centering
\setlength{\tabcolsep}{10pt}
\begin{tabular}{l r rr}
\toprule
\multirow{2}{*}{\bf Model} & \multirow{2}{*}{\bf Size} & \multicolumn{2}{c}{\bf BLEU}\\
\cmidrule(lr){3-4}
 & & Raw & KD \\
\midrule
MaskT                                         &  69M & 27.2 & 31.3 \\
~~ + Scaling                                  & 831M & 31.7 & 34.2 \\
\midrule
GLAT                                          &  71M & 24.0 & 30.6 \\
~~ + Scaling                                  & 835M & 31.4 & 33.4 \\
\bottomrule
\end{tabular}
\caption{Translation performance of large NAT models on WMT20 En$\rightarrow$De task. The size of NAT-Large model is (54, 6)$\times$1024.}
\label{tab:large}
\end{table}

\paragraph{Deeper Scaling}
\label{app:deeper_scaling}

To further explore the characteristics of depth scaling for NAT models, we deepen the encoder and decoder to 54-54 layers.
Results on WMT20 En$\rightarrow$De are shown in Table~\ref{tab:deeper-54}. Compared with the experimental results in Table~\ref{tab:deeper}, deeper NAT models (from (24,24) to (54,54)) improve higher performance on raw data than KD data (MasktT: +1.2 vs. +0.0; GLAT: +0.9 vs. +0.3). Encouragingly, depth scaling on raw data outperforms standard models trained on distillation data.

\paragraph{Component-Level Deeper Scaling}
\label{app:deeper_component}

To further verify the symmetry of NAT architecture, we conduct experiments on component-level depth scaling. 
Experimental results on WMT20 En$\rightarrow$De are shown in Table~\ref{tab:scale_component_deeper-54}. 
Regarding model type (fully or iterative NAT) and data type (raw or KD), performance gap between depth scaling encoder and depth scaling decoder is significant and stable. It indicates that scaling encoder is more important for NAT model than scaling decoder, which still holds for width scaling in Table~\ref{tab:scale_component}. Besides, comparing the depth scaling in Table~\ref{tab:deeper} and the deep encoder in Table~\ref{tab:scale_component_deeper-54}, NAT models with deep encoder show better performance than symmetric deep NAT ones.

\subsection{Results of Larger NAT Models}
\label{sec:larger_model}
In order to explore the upper-bound of translation performance for NAT, we enlarge models with both depth and width scaling. The model sizes are increased to 831M (MaskT) and 835M (GLAT).
Results are listed in Table~\ref{tab:large}. To the best of our knowledge, 34.2 BLEU could be a SOTA performance among existing NAT models. Comparing the performance of base NAT model on KD data and that of large NAT model on raw data, scaling may be an alternative way to replace knowledge distillation (MaskT: 31.7 vs. 31.3 and GLAT: 31.4 vs. 30.6).

\subsection{Results of Probing Tasks}
\label{sec:probing_task_for_raw}
To compare the scaling behaviors of AT and NAT models further, more experiments of probing task are conducted. The representations come from the NMT models trained on WMT20 En$\rightarrow$De raw data and the experimental results are depicted in Table~\ref{tab:probingraw}. The differences of scaling behaviors between AT and NAT models on raw data are similar to that on KD data in Table~\ref{tab:probing}.

\begin{table}
\centering
\setlength{\tabcolsep}{4.1pt}
\scalebox{0.95}{
\begin{tabular}{l r rrr rrr}
\toprule
\multicolumn{2}{c}{\multirow{2}{*}{\bf Task}} & \multicolumn{3}{c}{\bf{MaskT}} & \multicolumn{3}{c}{\bf{AT}} \\
\cmidrule(lr){3-5}\cmidrule(lr){6-8}
& & \bf{Base} & \bf{Big} & $\Delta$ & \bf{Base} & \bf{Big} & $\Delta$ \\
\midrule
\multirow{2}{*}{\rotatebox[origin=c]{90}{\small Surface}} & SeLen & 81.4 & 87.3 & +5.9 & 81.2 & 87.7 & +6.5 \\
& WC & 76.6 & 70.5 & -6.1 & 55.1 & 70.3 & +15.2 \\
\midrule
\multirow{3}{*}{\rotatebox[origin=c]{90}{\small Syntactic}} & TrDep & 34.6 & 35.9 & +1.3 & 35.6 & 36.5 & +0.9 \\
& ToCo  & 70.7 & 73.0 & +2.3 & 69.5 & 73.7 & +4.2 \\
& Bshif & 49.2 & 49.6 & +0.4 & 49.5 & 49.7 & +0.2 \\
\midrule
\multirow{5}{*}{\rotatebox[origin=c]{90}{\small Semantic}} & Tense & 82.6 & 84.1 & +1.5 & 83.5 & 85.0 & +1.5 \\
& SubN  & 79.3 & 80.6 & +1.3 & 79.3 & 81.8 & +2.5 \\
& ObjN  & 81.2 & 81.1 & -0.1 & 80.5 & 82.3 & +1.8 \\
& SoMo  & 49.8 & 49.9 & +0.1 & 49.9 & 49.9 & 0 \\
& CoIn  & 53.6 & 54.1 & +0.5 & 52.8 & 53.5 & +0.7 \\
\bottomrule
\end{tabular}}
\caption{Performance (accuracy $\uparrow$) of probing tasks for evaluating linguistic properties embedded in the learned representations of AT and NAT models (with width scaling) on \bf{raw data}.}
\label{tab:probingraw}
\end{table}

\subsection{Evaluation with METEOR}
\label{app:meteor}
To make the results convincing, another evaluation metric, METEOR, is used to measure the scaling behavior of NAT models. Different from BLEU, METEOR incorporates semantic information by calculating either exact match, stem match, or synonymy match. The results of METEOR are calculated with Multeval.\footnote{\url{https://github.com/jhclark/multeval}.} More experimental results of METEOR are provided in Table~\ref{tab:meteor}, which are similar to the results of BLEU in Table~\ref{tab:benchmark}.

\begin{table}[t]
\centering
\scalebox{0.94}{
\setlength{\tabcolsep}{2pt}
\begin{tabular}{l c rr rr}
\toprule
\multirow{2}{*}{\bf Model} & \textbf{W16 (0.6M)} & \multicolumn{2}{c}{\textbf{W14 (4.5M)} } & \multicolumn{2}{c}{\textbf{W20 (45.1M)} }\\
\cmidrule(lr){2-2}\cmidrule(lr){3-4}\cmidrule(lr){5-6}
& \textbf{Ro-En} & \textbf{En-De} &\textbf{De-En} & \textbf{En-De} &\textbf{De-En}\\
\midrule
\multicolumn{6}{c}{\bf MaskT} \\
Base  & 31.5 & 40.4 & 29.7 & 45.1 & 34.5 \\
Big   & 31.5 & 40.9 & \bf 29.8 & 46.3 & 34.9 \\
\hdashline
Cone  & \bf 31.7 & \bf 41.0 & \bf 29.8 & \bf 46.7 & \bf 35.0 \\
\midrule
\multicolumn{6}{c}{\bf GLAT} \\
Base   & 31.0 & 39.3 & 29.2 & 44.4 & 34.0 \\
Big    & 31.6 & 40.2 & 29.6 & 45.9 & \bf 34.5 \\
\hdashline
Cone   & \bf 31.7 & \bf 40.7 & \bf 29.7 & \bf 46.2 & \bf 34.5 \\
\bottomrule
\end{tabular}}
\caption{Translation performance (METEOR $\uparrow$) of the proposed NAT models on translation tasks with different data sizes. ``Cone'' denotes scaling NAT architecture to (12$\times$1024, 3$\times$256). The results of WMT16 En$\rightarrow$Ro are unavailable due to the lack of alignment information on Romanian for METEOR.}
\label{tab:meteor}
\end{table}

\begin{table}[t]
    \centering
    \scalebox{0.95}{
    \begin{tabular}{l m{5.0cm}}
    \toprule
    \bf Source & Sechs Maschinengewehre des Typs MG3 sind nach wie vor verschwunden. \\
    \hdashline
    \bf Refer. & Six MG3 machine guns are still missing. \\
    \midrule
    \bf GLAT-Base & Six MG3-\textcolor{red}{machine machine} guns have still disappeared.\\
    \hdashline
    \bf GLAT-Big & Six MG3 machine guns have still been missing.\\
    \bottomrule
    \end{tabular}}
    \caption{Examples about repetition for NAT models. Repetitive tokens are highlighted in \textcolor{red}{red color}.}
    \label{tab:repetition}
\end{table}

\begin{table}[t]
\scalebox{0.95}{
    \centering
    \begin{tabular}{l m{5.0cm}}
    \toprule
    \bf Source & \begin{CJK}{UTF8}{gbsn}国庆 长假 临近，\textcolor{red}{人们的 假期 计划 也 逐渐 敲定。}\end{CJK}\\
    \hdashline
    \bf Refer. & As the National Day holiday approaches, \textcolor{red}{people's holiday plans are gradually being finalized.}\\
    \midrule
    \bf GLAT-Base & The National Day long holiday near, \textcolor{red}{people people's plans plans gradually gradually gradually.}\\
    \hdashline
    \bf GLAT-Big & The National Day holiday is approaching, \textcolor{red}{people's holiday plans are gradually worked out.}\\
    \bottomrule
    \end{tabular}}
    \caption{Examples about fluency for NAT models. The key spans are highlighted in \textcolor{red}{red color}.}
    \label{tab:fluency}
\end{table}

\begin{table}[t]
\scalebox{0.95}{
    \centering
    \begin{tabular}{l m{5.0cm}}
    \toprule
    \bf Source & \begin{CJK}{UTF8}{gbsn}\textcolor{red}{曼西内利} 当时 虽然 有 上小学，但 后来 没有 毕业。\end{CJK}\\
    \hdashline
    \bf Refer. & Although \textcolor{red}{Mancinelli} entered elementary school, he did not graduate.\\
    \midrule
    \bf GLAT-Base & \textcolor{red}{Manthinelli} attended primary school at the time but but did not graduate.\\
    \hdashline
    \bf GLAT-Big & \textcolor{red}{Mancinelli} went attended primary school at the time but did not not graduate.\\
    \bottomrule
    \end{tabular}}
    \caption{Examples about word accuracy for NAT models. The key tokens are highlighted in \textcolor{red}{red color}.}
    \label{tab:accuracy}
\end{table}

\subsection{Commonly-Cited Weaknesses of NAT}
\label{app:weakness}
In this paper, we study the commonly-cited weaknesses of NAT from the following three perspectives: 1) {\em multimodality} indicated by token repetition ratio; 2) {\em generation fluency} calculated by language model perplexity; 3) {\em translation adequacy} measured by word translation accuracy. 
To illustrate the effect of scaling on commonly-cited weaknesses of NAT, examples are listed in Table~\ref{tab:repetition}, Table~\ref{tab:fluency} and Table~\ref{tab:accuracy} respectively.

\subsection{Training of NAT models}
\label{app:training}
We adopt Transformer-Base/Big configurations for all NAT models: both encoder and decoder contain 6 layers with 8/16 attention heads, the hidden dimension is 512/1024, and the feedforward layer dimension is 2048/4096. We train all NAT models with a big batch size of 480K. We train MaskT, GLAT models for 300K steps.

We list the training budget in Table~\ref{tab:parameters}. More details about training hyper-parameters can be found in the training scripts of different NAT models.

\begin{table}[t!]
\centering
\begin{tabular}{l rr}
\toprule
\bf Model   & \bf Size & \bf GPU Hours \\
\midrule
AT-Base     & 69M & 352h\\
AT-Big      &226M & 616h\\
\midrule
MaskT-Base  & 69M & 320h\\
MaskT-Big   & 226M& 584h\\
\hdashline
GLAT-Base   & 71M & 816h\\
GLAT-Big    &230M & 1120h\\
\bottomrule
\end{tabular}
\caption{The number of parameters and training budget (in GPU hours with 8 A100 for 300K steps). More detailed training hyper-parameters can be found in the training scripts of the different NAT models.}
\label{tab:parameters}
\end{table} 

\end{document}